%% file: paper.tex
\documentclass[sigconf]{acmart}
\graphicspath{{fig/}}

\usepackage[utf8]{inputenc}
\usepackage{url}
\usepackage{amsmath,bm,mathtools}
\usepackage{enumitem}
\usepackage{siunitx}
\usepackage{balance} 
\usepackage{subcaption}
\usepackage{algorithm}
\usepackage{algpseudocode}
\usepackage{xcolor}

\DeclareMathOperator*{\Argmax}{arg\,max}
\DeclareMathOperator*{\Argmin}{arg\,min}
\newlist{enuminline}{enumerate*}{1}
\setlist[enuminline,1]{label=\itshape\alph*\upshape)}

\copyrightyear{2023}
\acmYear{2023}
\setcopyright{acmlicensed}
\acmConference[KDD '23] {Proceedings of the 29th ACM SIGKDD Conference on Knowledge Discovery and Data Mining}{August 6--10, 2023}{Long Beach, CA, USA.}
\acmBooktitle{Proceedings of the 29th ACM SIGKDD Conference on Knowledge Discovery and Data Mining (KDD '23), August 6--10, 2023, Long Beach, CA, USA}
\acmPrice{15.00}
\acmDOI{10.1145/3580305.3599386}
\acmISBN{979-8-4007-0103-0/23/08}

\begin{document}

\title[Optimizing Recommendations for the Long-Term Without Delay]{Impatient Bandits: Optimizing Recommendations \\ for the Long-Term Without Delay}

\author{Thomas M. McDonald}
\authornote{This work was completed as part of an internship at Spotify.}
\orcid{0000-0001-7301-4399}
\email{tommcdonald955@gmail.com}
\affiliation{%
  \institution{University of Manchester}
  \country{United Kingdom}
}

\author{Lucas Maystre}
\orcid{0000-0002-8307-7673}
\email{lucasm@spotify.com}
\affiliation{%
  \institution{Spotify}
  \country{United Kingdom}
}

\author{Mounia Lalmas}
\orcid{0000-0002-3531-3096}
\email{mounia@acm.org}
\affiliation{%
  \institution{Spotify}
  \country{United Kingdom}
}

\author{Daniel Russo}
\orcid{0000-0001-5926-8624}
\email{djr2174@gsb.columbia.edu}
\affiliation{%
  \institution{Columbia University \& Spotify}
  \country{United States}
}

\author{Kamil Ciosek}
\orcid{0000-0002-0238-9393}
\email{kamilc@spotify.com}
\affiliation{%
  \institution{Spotify}
  \country{United Kingdom}
}

\renewcommand{\shortauthors}{Thomas M. McDonald, Lucas Maystre, Mounia Lalmas, Daniel Russo, \& Kamil Ciosek}

\begin{abstract}
\input{00-abstract}
\end{abstract}



\maketitle

\input{01-intro}
\input{02-relwork}
\input{03-method}
\input{04-podcasts}
\input{05-conclusion}

\bibliographystyle{ACM-Reference-Format}
\balance
\bibliography{paper}

\clearpage
\appendix
\input{0A-model}
\input{0B-covar}
\input{0C-contextual}

\end{document}

%% file: 00-abstract.tex
Recommender systems are a ubiquitous feature of online platforms.
Increasingly, they are explicitly tasked with increasing users' long-term satisfaction.
In this context, we study a content exploration task, which we formalize as a multi-armed bandit problem with delayed rewards.
We observe that there is an apparent trade-off in choosing the learning signal:
Waiting for the full reward to become available might take several weeks, hurting the rate at which learning happens, whereas measuring short-term proxy rewards reflects the actual long-term goal only imperfectly.
We address this challenge in two steps.
First, we develop a predictive model of delayed rewards that incorporates all information obtained to date.
Full observations as well as partial (short or medium-term) outcomes are combined through a Bayesian filter to obtain a probabilistic belief.
Second, we devise a bandit algorithm that takes advantage of this new predictive model.
The algorithm quickly learns to identify content aligned with long-term success by carefully balancing exploration and exploitation.
We apply our approach to a podcast recommendation problem, where we seek to identify shows that users engage with repeatedly over two months.
We empirically validate that our approach results in substantially better performance compared to approaches that either optimize for short-term proxies, or wait for the long-term outcome to be fully realized.

%% file: 01-intro.tex
\section{Introduction}
\label{sec:intro}

Many online platforms rely on recommender systems to assist users in finding relevant items among vast collections of content~\citep{ricci2015recommender}.
Applications are wide-ranging: recommender systems help individuals find books, movies or audio content~\citep{bennett2007netflix, mcinerney2018explore};
they help doctors find medical treatments for their patients~\citep{tran2021recommender}, and students find learning resources~\citep{verbert2012context}, among many others.
A key question underpins the design of any recommender system: What is a successful recommendation?
Across many applications, there is an ongoing shift towards defining success at longer time-horizons~\citep{zheng2018drn, zou2019reinforcement}, as long-term metrics are often better suited to capture users' satisfaction and platforms' goals~\citep{hohnhold2015focusing}.
For example, e-commerce platforms may want to maximize long-term revenue, subscription-based services may want to increase retention, and social platforms may want to encourage habitual engagement measured over several weeks or months.
In the context of podcast recommendations on an online audio streaming platform, recent work has shown that explicitly optimizing for long-term engagement (measured over a 60-day window post-recommendation) can significantly improve the user experience \citep{maystre2023optimizing}.
Most of the literature, however, implicitly assumes that there is sufficient data to estimate the long-term impact of recommendations.

\subsection{Content Exploration Problem}

In this paper, we focus on a specific aspect of recommender systems and seek to address a content exploration problem.
On most online platforms, new content is released regularly.
In order to learn about that content's appeal, we must first recommend it to users.
This is known as the \emph{cold-start problem}.
After ensuring an adequate amount of information has been gathered, an effective system should rapidly shift recommendations away from poor content.

\begin{figure}[t]
  \centering
  \includegraphics[width=0.8\linewidth]{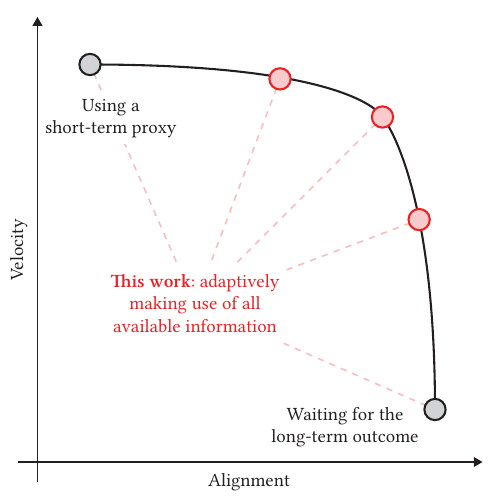}
  \caption{Short-term proxies enable a rapid feedback loop, but might be poorly aligned with long-term success metrics, which take longer to realize.
  Our method finds the optimal tradeoff by adaptively making use of all available information at a given time.}
  \label{fig:tradeoff}
\end{figure}

We formalize this task as a \emph{multi-armed bandit} problem, where we seek to identify promising content through successive interactions with users \citep{li2010contextual}.
Optimizing for long-term definitions of success in the bandit setting is challenging, as long-term metrics are---by construction---delayed~\citep{hong2019tutorial}.
This gives rise to an apparent tradeoff, illustrated in Figure~\ref{fig:tradeoff}, between using short-term proxies that are observable quickly (top-left) and ensuring that actions selected are aligned with long-term success (bottom-right).
We propose a means of circumventing this tradeoff by exploiting the insight that \emph{most long-term outcomes become increasingly predictable over time}.

Driven by practical applications, we assume that intermediate outcomes are progressively revealed over time, from the moment the action is selected up to the moment the full reward is observed.
We call this the \emph{progressive feedback} setting.
We develop a probabilistic model that forms beliefs about the delayed rewards an arm generates on the basis of outcomes observed so far.
As time passes, uncertainty diminishes and the model is able to make increasingly precise predictions.
To facilitate this, we contend that historical data from distinct but similar applications (e.g., previous content releases) can be used to learn the association between intermediate and long-term outcomes.
In effect, we propose a meta-learning approach that learns to infer long-term outcomes of interest from intermediate observations, revealed progressively over time.
We then take advantage of this reward model to address our sequential decision-making problem, by combining the predictive model with a bandit algorithm.
The bandit uses probabilistic predictions from the model to efficiently balance exploration and exploitation.
Even if the first few intermediate outcomes are insufficient to perfectly infer the average delayed reward an arm generates, they might be sufficient to reveal that this arm is outperformed by others.
In such cases, our bandit algorithm will shift effort away from the arm.
Note that, in contrast to well-studied bandit settings where feedback is observed at once, either immediately or after a given delay, the progressive feedback setting presents distinctive challenges:
Information can be obtained actively, by selecting an action, or passively by letting time unfold and incrementally receiving new data about the outcomes of actions taken in the past.

Our methodology is very general, and can be applied to a wide range of problems.
In this work, we consider a recently-studied podcast recommendation application~\citep{maystre2023optimizing}.
In this application, actions correspond to podcast shows, and the reward is defined as the number of days a user engages with a show in the 59 days that follow a successful recommendation.
Intermediate outcomes consist of binary activity indicators for each of the 59 days, observed with the corresponding delay.
We evaluate our approach using data from the Spotify audio streaming platform, and show that \begin{enuminline}
\item the full reward can be accurately predicted after only a few days of observation, and that
\item the content-exploration problem can be solved much quicker than approaches that rely on short-term proxies or wait for the full reward to become available.
\end{enuminline}

\paragraph{Summary of Contributions}
In this work, we make the following contributions.
\begin{itemize}
    \item A Bayesian filtering approach to reward estimation, which enables us to incorporate all available information in order to predict delayed outcomes and quantify uncertainty (Section~\ref{sec:model}).
    \item A meta-learning approach, where the prior and noise covariance structures that power Bayesian filtering are themselves learned from data. The method learns across items how to make rapid inferences about a new item (Section~\ref{sec:training}).
    \item The \emph{impatient bandit algorithm}, a novel algorithm for the progressive feedback setting, which uses intermediate information received at each round to iteratively update the Bayesian filter, and enables us to efficiently balance exploration and exploitation whilst providing recommendations that optimize for long-term engagement (Section~\ref{sec:bandit}).
    \item An application of our impatient bandit algorithm to a real-world podcast recommendation problem, presented alongside empirical results, that show that our proposed method considerably outperforms approaches based on fully-delayed feedback or short-term proxy metrics (Section~\ref{sec:podcasts}).
\end{itemize}


%% file: 02-relwork.tex
\section{Related Work}
\label{sec:relwork}

We start by briefly discussing relevant related work on multi-armed bandits and applications to recommender systems.

\paragraph{Multi-Armed Bandits}
Often used to model online platforms~\citep{mattos2019multi}, multi-armed bandits (MAB) formalize a simple sequential decision-making problem, where at each round $t$ an agent selects one of several possible actions and receives a corresponding reward $r_t$.
The goal is usually to maximize the sum of rewards received over a given time horizon.
The simplest and most widely studied bandit setting is the \emph{strictly sequential feedback} scenario, where $r_t$ is immediately observed~\citep{srinivas2009gaussian, agrawal2012analysis}.
However many extensions have been proposed. 

One such extension is the case of \emph{parallelized actions}.
Rather than simply selecting a single action at each round and receiving a single corresponding reward, we can also consider a scenario in which multiple actions can be taken at each round in parallel~\citep{desautels2014parallelizing, kandasamy2018parallelised}.
This extension is also referred to as the \emph{batched} bandit setting.
The challenge arises from having to concurrently select several actions without knowing the reward associated with the other actions in the batch.

An additional variant of the MAB relevant to our work is the \emph{delayed feedback} setting, which can be viewed as a generalization of the batch feedback setting \citep{kandasamy2018parallelised}.
In this setting, the reward $r_t$ is only revealed after a delay of $\Delta$ rounds, i.e., at round $t + \Delta$.
In this case, the agent is forced to make a series of decisions without knowledge of the results of all previous actions taken.
Prior work in this area has utilized Thompson sampling~\citep{chapelle2011empirical, kandasamy2018parallelised} and upper-confidence bound (UCB)~\citep{joulani2013online, desautels2014parallelizing} algorithms to address this setting.

\paragraph{Thompson Sampling}
Thompson sampling is a class of algorithms used for sequential decision-making in bandit settings, that efficiently balances \emph{exploration} of the action space with \emph{exploitation} of actions that are believed to be associated with comparatively large rewards~\citep{russo2018tutorial}.
Whilst the technique was introduced almost a century ago~\citep{thompson1933likelihood}, it has become increasingly popular over the course of the last decade due to its strong empirical performance when applied to modern, large-scale online learning problems~\citep{scott2010modern, graepel2010web}.

In the face of uncertainty, Thompson sampling randomizes among all actions that are plausibly optimal.
This allows for greater robustness to delayed feedback compared to algorithms based on upper-confidence bounds, which are deterministic and rely on rapidly changing beliefs to adjust which arm is sampled.
There is both theoretical and empirical evidence that, due to its randomized nature, Thompson sampling is resilient to delayed feedback~\citep{chapelle2011empirical, kandasamy2018parallelised, qin2022adaptivity, wu2022thompson}.
This characteristic of the algorithm is crucial in our problem.

In our work, we use Thompson sampling in combination with a belief model that predicts a long-term metric using progressively revealed intermediate observations.
We meta-learn this model on historical data.
This aspect of our work connects to a series of recent papers that develop provable bounds on the loss in performance due to fitting the prior used in Bayesian bandit algorithms from past data~\citep{bastani2022meta, simchowitz2021bayesian, basu2021no}.

\paragraph{Intermediate Feedback}
Several recent papers have employed Thompson sampling in settings with delayed outcomes, but useful intermediate observations~\citep{yang2020targeting, caria2020adaptive, wu2022partial}.
UCB algorithms have also been applied to this setting by~\citet{grover2018best}, who consider a scenario where noisy observations of the true feedback are received at intermediate rounds between $t$ and $t+\Delta$.
Key differences between their work and ours include the fact that \citeauthor{grover2018best} consider the problem of top-$k$ best-arm identification with a stochastic delay, in contrast to our objective of cumulative regret minimization with a fixed delay.
Another differentiating factor of greater consequence is the fact that they assume that the intermediate feedback consists of independent random variables, whereas our progressive feedback is crucially not i.i.d., which is how our model is able to effectively generalize.
Additionally, our work employs Bayesian filtering to seamlessly perform inference, an approach explicitly motivated by a real use-case, where historical data allows us to fit an informed prior.

Outside of the literature on MABs, \citet{prentice1989surrogate} and \citet{athey2019surrogate} formalize conditions under which intermediate feedback can be used to estimate long-term outcomes.
They also find empirically that intermediate feedback can lead to both increased accuracy and precision in estimates of long-term outcomes.

\paragraph{Recommender Systems \& Long-Term Goals}
Bandits are a popular approach for addressing many types of recommendation problem.
In a seminal paper, \citet{li2010contextual} use a contextual variant of the bandit problem to personalize recommendations on a news platform, and more recently, \citet{aziz2022identifying} use MABs to recommend podcasts by maximizing the impression-to-stream rate.

Optimizing recommendations for long-term user engagement is a problem that is of great practical interest in industry, and bandits have also been used previously to address this specific scenario.
\citet{wu2017returning} address this problem using a UCB algorithm that models the temporal return behaviour of users to maximize the cumulative number of clicks from a group of users over a period of time.

Beyond bandits, more general reinforcement learning (RL) approaches have also been applied to the problem of maximizing long-term user engagement~\citep{zheng2018drn, zou2019reinforcement}.
``Full'' RL enables principled reasoning about inter-temporal tradeoffs and delayed rewards, at the expense of increased complexity.
Implementing effective RL algorithms that address realistic recommender system problems is non-trivial due to the challenging nature of off-policy learning and evaluation~\citep{zhao2019deep}.

%% file: 03-method.tex
\section{Methodology}
\label{sec:method}

We present our approach to solving the content exploration problem outlined in the introduction.
We adopt the terminology of multi-armed bandits.
We consider a set of $N$ actions, $\mathcal{A} = \{a_1, \ldots, a_N\}$, corresponding, e.g., to different recommendation candidates.
At each round $t = 1, 2, \ldots$, we select one or more actions.
For every action we select, we observe a reward $r_a$ after a delay of $\Delta$ rounds, i.e., at round $t + \Delta$.
Informally, we seek to develop a methodology that helps us quickly identify and exploit actions with high mean reward $\bar{r}_a = \mathbb{E}[r_a]$.
We assume that the reward $r_a$ is a function of intermediate observations $z_{a,1}, \ldots, z_{a,K}$, that become available progressively during the interval $[t, t + \Delta]$ after selecting the action.
We call this the \emph{progressive feedback} setting.

In Section~\ref{sec:model}, we consider a fixed action $a$ and develop a Bayesian reward model that takes advantage of intermediate observations to estimate the mean reward $\bar{r}_a$.
In Section~\ref{sec:training}, we take advantage of historical data to estimate the parameters of the reward model, effectively instantiating a \emph{meta-learning} approach. 
Building on this model, in Section~\ref{sec:bandit}, we develop a bandit algorithm that efficiently balances exploration and exploitation in the progressive feedback setting.

\paragraph{Concrete Example}
While this section introduces the methodology in a generic way, it is helpful to keep a concrete application in mind.
In Section~\ref{sec:podcasts} we consider a podcast recommendation problem, where the actions $\mathcal{A}$ correspond to podcast shows.
The reward $r$ is the cumulative engagement with a podcast show over a period of $\Delta$ days: $r_a = \sum_{i=i}^\Delta z_{a,i}$.

\subsection{Bayesian Reward Model}
\label{sec:model}

We consider a fixed action $a$ and, for conciseness, we omit $a$ from all subscripts.
Let $r$ be the sample reward and $\bar{r} = \mathbb{E}[r]$ be the mean reward associated to selecting the action.
Define the sample trace, $\bm{z} = (z_1, \ldots, z_K) \in \mathbf{R}^K$, as a vector containing intermediate outcomes.
We assume that $z_k$ is observed after $\Delta_k \le \Delta$ rounds, and, 
without loss of generality, that $\Delta_1 \le \cdots \le \Delta_K$.
Correspondingly, we define the average trace as $\bar{\bm{z}} = \mathbb{E}[\bm{z}]$.
We postulate the following generative model of sample traces $\{\bm{z}_m\}$:
\begin{align}
\label{eq:genmodel}
\bar{\bm{z}} &\sim \mathcal{N}(\bm{\mu}, \bm{\Sigma}),
&\bm{z}_m  &= \bar{\bm{z}} + \bm{\varepsilon}_m,
&\bm{\varepsilon}_m &\sim \mathcal{N}(\bm{0}, \bm{V}) \ \text{i.i.d.}
\end{align}
That is, we assume a priori that the average trace $\bar{\bm{z}}$ corresponding to the action is sampled from a multivariate Gaussian distribution with mean $\bm{\mu}$ and covariance matrix $\bm{\Sigma}$, and that
a sample trace $\bm{z}_m$ is a noisy copy of $\bar{\bm{z}}$, corrupted by additive zero-mean Gaussian noise with covariance matrix $\bm{V}$, independently for each $m$.
Furthermore, we assume that we can reconstruct the reward from all intermediate observations as
\begin{align*}
r = \bm{w}^\top \bm{z},
\end{align*}
where $\bm{w} \in \mathbf{R}^K$ is a vector of weights.
By the linearity of expectation, it follows that $\bar{r} = \bm{w}^\top \bar{\bm{z}}$.
We treat $\bm{w}$ as given, and $\{\bm{\mu}, \bm{\Sigma}, \bm{V} \}$ as model parameters.
We discuss how to learn them from data in Section~\ref{sec:training}.

Assume that we are at round $t$ and that we have selected the action $M$ times so far, at rounds $t_1 \le \cdots \le t_m \le t$.
We represent the observations collected at round $t$ as a dataset of $M$ independent traces, $\mathcal{D} = \{(\bm{z}_m, \ell_m) : m = 1, \ldots, M\}$.
Some traces might only be partially observed, and we use $\ell_m \doteq \max \{ k : \Delta_k \le t - t_m \}$ to index the last element of $\bm{z}_m$ that is observed at round $t$.

\subsubsection{Iterative Belief Updates}

We consider the problem of estimating $\bar{r}$ given $\mathcal{D}$.
Instead of reasoning about $\bar{r}$ directly, we begin by addressing the problem of estimating $\bar{\bm{z}}$.
We take a Bayesian approach and seek to compute the posterior distribution
\begin{align*}
p(\bar{\bm{z}} \mid \mathcal{D}) \propto p(\mathcal{D} \mid \bar{\bm{z}}) \mathcal{N}(\bar{\bm{z}} \mid \bm{\mu}, \bm{\Sigma}).
\end{align*}
Given our generative model~\eqref{eq:genmodel}, we will show that the posterior remains Gaussian, even in the presence of partially observed traces.
Finally, writing $p(\bar{\bm{z}} \mid \mathcal{D}) \doteq \mathcal{N}(\bar{\bm{z}} \mid \bm{\mu}', \bm{\Sigma}')$ and given that $\bar{r}$ is a linear function of the mean trace $\bar{\bm{z}}$, we have that
\begin{align*}
\bar{r} \mid \mathcal{D} \sim \mathcal{N}(\mu, \sigma^2),
\end{align*}
where $\mu = \bm{w}^\top \bm{\mu}'$ and $\sigma^2 = \bm{w}^\top \bm{\Sigma}' \bm{w}$.

We describe the process by which we fold in a single trace into the belief. 
The full posterior can be obtained by repeating this procedure iteratively, $M$ times.
For conciseness, we drop the subscript $m$ and denote the trace and cutoff index as $(z, \ell)$, respectively.
We denote by $\bm{A}_{:i, :j}$ the submatrix obtained by taking the $i$ first rows and the $j$ first columns of a matrix $\bm{A}$.
Similarly, we denote by $\bm{a}_{:i}$ the first $i$ elements of the vector $\bm{a}$.
Thanks to the self-conjugacy property of the Gaussian distribution, we can write the posterior distribution of $\bar{\bm{z}}$ after observing the $\ell$ first elements of the trace $\bm{z}$ as a multivariate Gaussian with mean vector and covariance matrix
\begin{align*}
\bm{\mu}' &=
    \bm{\mu} + \bm{\Sigma}_{:K, :\ell} \left(\bm{\Sigma}_{:\ell, :\ell} + \bm{V}_{:\ell, :\ell} \right)^{-1} \left(\bm{z}_{:\ell} - \bm{\mu}_{:\ell} \right), \\
\bm{\Sigma}' &=
    \bm{\Sigma} + \bm{\Sigma}'_{:K, :\ell} \left(\bm{\Sigma}_{:\ell, :\ell} + \bm{V}_{:\ell, :\ell} \right)^{-1} \bm{\Sigma}_{:\ell, :K},
\end{align*}
respectively.
We refer the reader to~\citet[Section A.2]{rasmussen2006gaussian} for more details on these update equations.
The complete iterative procedure is provided in Algorithm~\ref{alg:inference}.

\begin{algorithm}[t]
\caption{Computing the posterior of $\bar{z}$.}
\label{alg:inference}
\begin{algorithmic}[1]
\Require Parameters $\bm{\mu}, \bm{\Sigma}, \bm{V}$, dataset $\mathcal{D}$
\State $\bm{\mu}' \gets \bm{\mu}$
\State $\bm{\Sigma}' \gets \bm{\Sigma}$
\For{$(\bm{z}, \ell) \in \mathcal{D}$}
\State $\bm{A} \gets \bm{\Sigma}'_{:K, :\ell} (\bm{\Sigma}'_{:\ell, :\ell} + \bm{V}_{:\ell, :\ell} )^{-1}$
\State $\bm{\mu}' \gets
    \bm{\mu} + \bm{A} \left(\bm{z}_{:\ell} - \bm{\mu}'_{:\ell} \right)$
\State $\bm{\Sigma}' \gets
    \bm{\Sigma}' + \bm{A} \bm{\Sigma}'_{:\ell, :K}$
\EndFor
\end{algorithmic}
\end{algorithm}

\paragraph{A Note on Gaussian Noise}
The assumption in~\eqref{eq:genmodel} that each trace $\bm{z}$ is Gaussian with mean $\bar{\bm{z}}$ might seem restrictive at first sight.
For example, in Section~\ref{sec:podcasts}, we consider binary observation vectors $\bm{z} \in \{0, 1\}^\Delta$, for which a Gaussian is arguably a poor model.
In fact, given that our ultimate goal is to infer $\bar{\bm{z}}$ from several traces, the impact of this assumption is relatively benign.
To see this, assume that we are given $M$ full traces $\bm{z}_1, \ldots, \bm{z}_M$ such that $z_m = \bar{\bm{z}} + \bm{\varepsilon}_m$, where $\{\bm{\varepsilon}_m\}$ are independently and identically distributed but not necessarily Gaussian.
It can be shown that the empirical average $\hat{\bm{z}} = M^{-1} \sum_m \bm{z}_m$ is a sufficient statistic for $\bar{\bm{z}}$ given $\{ \bm{z}_m \}$.
For $M$ large, we can invoke the central limit theorem to argue that a Gaussian approximation for $\hat{\bm{z}}$ (and, correspondingly, a Gaussian approximation for the individual traces $\bm{z}_1, \ldots, \bm{z}_M$) is accurate for the purpose of estimating $\bar{\bm{z}}$. 

\paragraph{Optimizing the Implementation}
For simplicity, we have described Bayesian inference in our model as a sequential procedure.
In practice, there are several ways in which Algorithm~\ref{alg:inference} can be made more computationally efficient.
These include \begin{enuminline}
\item updating the posterior using multiple traces in a single batch, instead of processing each trace independently;
\item performing incremental updates by reusing beliefs from previous rounds; and
\item only updating beliefs for actions that have received new observations.
\end{enuminline}

\subsection{Training the Reward Model}
\label{sec:training}

A crucial aspect of our method is the ability to take advantage of past data to learn the model parameters $\{\bm{\mu}, \bm{\Sigma}, \bm{V}\}$.
Specifically, we assume access to historical data about a different set of actions $\mathcal{A}'$.
In the context of a recommender system, for example, this could be existing content for which we already have a sufficient amount of interaction data.
For each $a \in \mathcal{A}'$, denote by $\mathcal{H}_a = \{ (\bm{z}_{am}, r_{am}) : m = 1, \ldots, M_a \}$ the data corresponding to action $a$.

For each action $a \in \mathcal{A}'$, we  begin by computing the empirical mean trace vector and noise covariance matrix
\begin{align*}
\hat{\bm{z}}_a &= M_a^{-1} \sum_{\bm{z} \in \mathcal{H}_a} \bm{z},
&\hat{\bm{V}}_a &= M_a^{-1} \sum_{\bm{z} \in \mathcal{H}_a} (\bm{z} - \hat{\bm{z}}_a)(\bm{z} - \hat{\bm{z}}_a)^\top,
\end{align*}
respectively.
We then estimate the model parameters $\bm{\mu}, \bm{\Sigma}, \bm{V}$ by using empirical averages, as
\begin{align*}
\bm{\mu}
    &= \lvert \mathcal{A}' \rvert^{-1} \sum_{a \in \mathcal{A}'} \hat{\bm{z}}_a, \\
\bm{\Sigma}
    &= \lvert \mathcal{A}' \rvert^{-1} \sum_{a \in \mathcal{A}'} (\bm{\mu} - \hat{\bm{z}}_a)(\bm{\mu} - \hat{\bm{z}}_a)^\top, \\
\bm{V}
    &= \lvert \mathcal{A}' \rvert^{-1} \sum_{a \in \mathcal{A}'} \hat{\bm{V}}_a.
\end{align*}
In principle, more advanced estimation methods might be used, such as type-II maximum likelihood, also known as \emph{empirical Bayes} \citep{rasmussen2006gaussian}.
In practice, however, we have found that the simple empirical averages described above are very effective.

Intuitively, the covariance matrices $\bm{\Sigma}$ and $\bm{V}$ play a critical role in our approach.
They encode the correlations between outcomes observed at different points in time.
If intermediate outcomes observed early on are highly predictive of later outcomes, we expect that we can accurately estimate $\bar{\bm{z}}$ (and thus $\bar{r}$) without waiting for the full $\Delta$ rounds required to observe $r$.
We will revisit this from an empirical perspective in Section~\ref{sec:evalmodel}.

\paragraph{A Note on the Weights}
Our approach assumes that the reward is a given linear function of the trace.
For example, in Section~\ref{sec:podcasts}, we consider a problem where the reward is defined as $r = \sum_k z_k$, corresponding to $\bm{w} \doteq \bm{1}$.
In practice, one might try to fit long-term objectives to a linear model, by solving a regression problem
\begin{align*}
\textstyle
\Argmin_{w} \sum_{a \in \mathcal{A}'} \sum_{(\bm{z}, y) \in \mathcal{H}_a'} (y - \bm{w}^\top \bm{z})^2,
\end{align*}
where $y$ is a target that is not exactly a linear function of $\bm{z}$.
In this case, it is important to note that the reward $r = \bm{w}^\top \bm{z}$ is an approximation of the true objective $y$.
We briefly elaborate on this in Appendix~\ref{app:model}.

\subsection{Bandit Algorithm}
\label{sec:bandit}

Equipped with a model capable of making inferences about the arms' mean rewards given intermediate observations, we can now develop a bandit algorithm that works effectively in the progressive-feedback setting, where information about the reward is revealed progressively over multiple rounds.

Although several different objectives for the bandit problem exist in the literature, in this work we focus on the goal of minimizing the \emph{cumulative expected regret}.
In the case of a single action being selected at each round, we define the cumulative expected regret at round $T$ as
\begin{align*}
\mathbb{E} \left[ R_T \right] = \mathbb{E}\left[\sum_{t=1}^T (\bar{r}^\star - \bar{r}_t)\right],
\end{align*}
where $\bar{r}^\star$ is the mean reward obtained by selecting the best action, $r_t$ is the mean reward corresponding to the action selected at round $t$, and the expectation is taken over the algorithm's internal randomization over actions \citep{slivkins2019introduction}.
We extend this definition to the case where we select multiple actions in parallel at each round, as
\begin{align*}
\mathbb{E} \left[ R_T \right] = \mathbb{E} \left[ \sum_{t=1}^T \left(\bar{r}^\star - B^{-1} \sum_{i = 1}^B \bar{r}_{t, i}\right) \right],
\end{align*}
where $B$ is the number of actions per round, and $\bar{r}_{t, i}$ is the mean reward associated to the $i$th action performed at round $t$.

Before describing our algorithm, we first present a brief overview of Thompson sampling~\citep{russo2016information,russo2018tutorial}. 
\citet{slivkins2019introduction} give a generalized formulation of Thompson sampling for bandits with immediately observable rewards, which we simplify here for ease of exposition.
In a strictly sequential multi-armed bandit, when an agent takes an action $a_t \in \mathcal{A}$, a corresponding reward $r_t \sim q_{\theta}(\cdot \mid a_t)$ is observed.
We place a prior distribution $p$ over the model parameters $\theta$.
The action to be taken at each round is chosen by computing $a_t \gets \Argmax_{a \in \mathcal{A}} \mathbb{E}_{q_{\hat{\theta}}} [r_t \mid a_t = a]$, yielding a realized observation, which we then condition on to update $p$.
Rather than taking a \emph{greedy} approach, whereby $\hat{\theta}$ is the expectation of $\theta$ with respect to $p$, Thompson sampling instead samples the parameters from $p$ (i.e. $\hat{\theta} \sim p$).
This is a subtle, but powerful difference, as it ensures that the algorithm does not purely exploit actions that yield large rewards in the first few rounds of feedback, ignoring other, possibly better actions.
Due to the non-zero variance of the belief on the mean reward associated with each action, Thompson sampling may select an action other than that which the greedy algorithm would deem optimal.
This mechanism trades off exploration and exploration effectively, and is known to achieve low cumulative regret~\citep{russo2018tutorial}.

\subsubsection{Impatient Bandit Algorithm}
Our approach builds on the Thompson sampling algorithm, applying it to the progressive feedback setting.
In our case, the parameters $\theta$ simply correspond to the average rewards $\{ \bar{r}_a : a \in \mathcal{A} \}$.
The key to our approach is to make use of the reward model developed in Section~\ref{sec:model} to infer beliefs $p(\bar{r}_a)$.
By updating beliefs based intermediate outcomes, we enable the sampling step in the Thompson sampling to take full advantage of \emph{all} information collected up to round $t$, and not only of fully observed rewards.
We call the resulting procedure the \emph{impatient bandit} and describe it in Algorithm~\ref{alg:bandit}.

\begin{algorithm}[t]
\caption{Impatient Bandit Algorithm}\label{alg:bandit}
\begin{algorithmic}[1]
\Require Actions $\mathcal{A}$, number of actions per round $B$
\For{$t = 1, \dots, T$}
  \For{$a \in \mathcal{A}$}
    \State Update $\mathcal{D}_a$ with new observations
    \State $p(\bar{\bm{z}}_a) \gets \mathcal{N}(\bar{\bm{z}}_a \mid \bm{\mu}_a, \bm{\Sigma}_a)$ via Algorithm~\ref{alg:inference} on $\mathcal{D}_a$
    \State $p(\bar{r}_a) \gets \mathcal{N}(\bar{r}_a \mid \bm{w}^\top \bm{\mu}_a, \bm{w}^\top \bm{\Sigma}_a \bm{w})$
  \EndFor
  \For{$i = 1, \dots, B$}
    \For{$a \in \mathcal{A}$}
      \State Sample mean reward $\hat{r}_a \sim p(\bar{r}_a)$ 
    \EndFor
    \State Take action $a_{t,i} \gets \Argmax_{a \in \mathcal{A}} \{ \hat{r}_a \}$
  \EndFor 
\EndFor
\end{algorithmic}
\end{algorithm}

%% file: 04-podcasts.tex
\section{Application to Podcasts}
\label{sec:podcasts}

\begin{figure*}[t]
  \centering
  \includegraphics{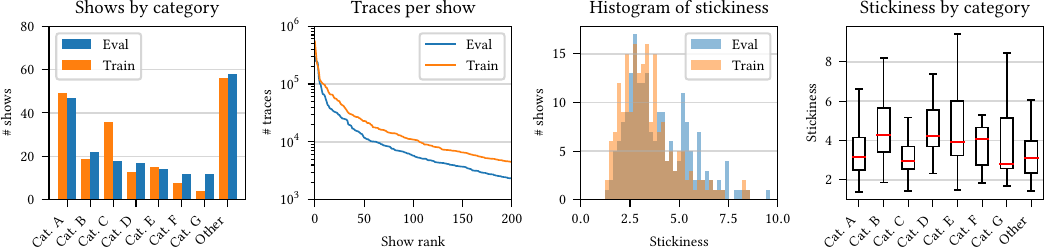}
  \caption{Summary statistics of a dataset of podcast shows and corresponding consumption traces.
  There is large heterogeneity in show-stickiness (center-right), even when controlling for category (boxplot, right).}
  \Description{Summary statistics of a dataset of podcast shows and corresponding consumption traces.
  There is large heterogeneity in show-stickiness (center-right), even when controlling for category (boxplot, right).}
  \label{fig:spotifydata}
\end{figure*}

We consider a concrete application of our content-exploration problem to podcast recommendations on Spotify, a leading online audio streaming platform.\footnote{%
See: \url{https://newsroom.spotify.com/company-info/}.}
In Section~\ref{sec:setup}, we begin by describing how our generic methodology can be applied to optimizing long-term user engagement with podcasts, and present a real-world dataset of podcast consumption traces.
In Section~\ref{sec:evalmodel}, we study the reward model in isolation, and evaluate its predictive accuracy.
In Section~\ref{sec:evalbandit}, we consider a sequential decision-making task in the progressive-feedback setting, and compare the empirical performance of our impatient bandit against competing approaches.

To complement the experiments presented in this section, we provide a companion software package with a reference implementation of our algorithm.\footnote{%
See: \url{https://github.com/spotify-research/impatient-bandits}.}
While we are unable to publicly release the data due to confidentiality reasons, our package includes a synthetic dataset that leads to comparable findings.

\subsection{Problem Formulation \& Data}
\label{sec:setup}

Traditionally, podcast recommender systems optimize for short-term rewards, such as the click-through-rate~\citep{aziz2022identifying}.
Recently, \citet{maystre2023optimizing} show that explicitly optimizing podcast recommendations for long-term outcomes can lead to substantial impact on a real-world, large-scale recommendation problem.
They propose a system that reasons simultaneously about the \emph{clickiness} (i.e., the click-through-rate) and the \emph{stickiness} of a recommendation.
Stickiness is defined in terms of the downstream consequences of a successful recommendation. In particular, the authors suggest counting the number of days users engage with a podcast show discovered through a recommendation over the \num{59} days that follows a first listen.
In this work, we adopt their definitions and optimization metrics, but consider a specific subset of the overall recommendation problem.
We focus on estimating stickiness (i.e., we do not model the click-through rate), and seek to quickly identify new podcast shows that have high average stickiness.
This lets us investigate the challenging problem of estimating long-term rewards for new content \emph{in isolation}, without being confounded by other aspects of the overall recommendation problem.
\citet{maystre2023optimizing} discuss how to estimate the click-through rate, and how to personalize models  to take into account users' preferences, but they do not address the content exploration problem we study here.

Formally, we instantiate the methodology described in Section~\ref{sec:method} as follows.
The set of actions $\mathcal{A}$ corresponds to $N$ candidate podcast shows that are new and that we need to explore.
We define the reward $r \in \{0, \ldots, 59 \}$ as the number of days a user engages with a show in the 59 days that follow a successful recommendation.\footnote{%
For the purposes of this paper, note that such a long horizon crystallizes the challenges of optimizing for the long-term, and forces us to develop methods that explicitly address these challenges.}
This reward is observed with a delay of $\Delta = 60$ days.
We refer to the mean reward $\bar{r}_a$ corresponding to show $a$ as the \emph{stickiness} of the show.
We collect intermediate outcomes $z_k = \mathbf{1}\{\text{the user engaged on day $k$}\}$ into an activity trace $\bm{z} \in \{0, 1\}^{59}$.
Naturally, each activity indicator $z_k$ is observed with delay $\Delta_k = k+1$.
From these definitions, it follows that $r = \sum_k z_k = \bm{w}^\top \bm{z}$, where $\bm{w} = \bm{1}$ is the all-ones vector.
The distinct set $\mathcal{A}'$ and historical data $\mathcal{H}_a$, $a \in \mathcal{A}'$ correspond to a set of established shows and the corresponding historical consumption traces, respectively.
We seek to develop a bandit algorithm that learns to maximize the long-term engagement attributable to each recommendation.
This is a clear instance of the progressive feedback setting;
Every day, actions must be taken with only partial knowledge about the outcome of decisions made in the previous 59 days.

\subsubsection{Dataset}
\label{sec:datadesc}

We consider a dataset of podcast consumption traces collected on the Spotify audio streaming platform between September 2021 and May 2022.
The data is divided into a training set and an independent validation set.
Each subset consists of a sample of \num{200} podcast shows first published on the platform during a given three-month period.
For each of these shows, the data contains a representative sample of users that discover the show during the same three-month period.
For each user, we obtain a longitudinal trace that captures their engagement with the show on each day starting from the day of discovery,\footnote{%
We define a discovery as the first stream that happens on the platform.} in the form of a $59$-dimensional binary vector.
The training and validation sets cover podcast shows appearing during the periods September--December 2021 and January--March 2022, respectively.
Each subset covers a distinct set of shows.

The podcast shows included in the dataset span a wide range of categories, from \emph{Arts} to \emph{True Crime}.
Figure~\ref{fig:spotifydata} (left) shows that the distribution of shows over categories is comparable across the two periods.\footnote{%
For confidentiality reasons, we obfuscate the names of the categories.}
In total, the dataset consists of \num{8.77}M activity traces, corresponding to a total of \num{26}M cumulative active-days.
The number of traces per show ranges between \num{2.4}K and \num{295}K, with a median of \num{5.8}K (Figure~\ref{fig:spotifydata}, center-left).
For each show, we define the ground-truth stickiness by means of the empirical average (across users) of the cumulative active-days.
Figure~\ref{fig:spotifydata} (center-right) shows that there is substantial heterogeneity in stickiness across shows, with the lower quartile, median, and upper quartile at \num{2.6}, \num{3.4}, and \num{4.6} days, respectively.
This suggests that the downstream impact of a discovery can be very different across shows.
We note that the stickiness histogram is comparable across the two subsets.
Finally, some categories appear to be somewhat stickier than others, but within-category variability is significantly larger than between-category variability (Figure~\ref{fig:spotifydata}, right).

\subsection{Evaluating the Reward Model}
\label{sec:evalmodel}

\begin{figure}[t]
  \centering
  \includegraphics{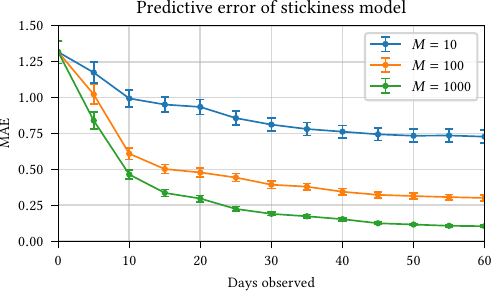}
  \caption{Mean absolute error ($\pm$ standard error) of stickiness estimates as a function of days of data observed.}
  \Description{A plot showing how the mean absolute error achieved by our Bayesian stickiness model varies as a function of how many days of activity data we have access to, visualized for scenarios in which we have access to 10, 100 and 1000 user traces.}
  \label{fig:prederror}
\end{figure}

We focus first on evaluating our Bayesian reward model in isolation.
We estimate $\bm{\mu}$, $\bm{\Sigma}$ and $\bm{V}$ by using the shows and consumption traces contained in the training dataset.
For each show in the validation dataset, we randomly sample 2000 user traces.
From this subset, we use $M$ traces to infer the stickiness of each show (via Algorithm~\ref{alg:inference}), and we use the remaining ($2000 - M$) traces for computing the ground truth empirical stickiness.
In Figure~\ref{fig:prederror}, we visualize how the predictive accuracy of our stickiness model varies as a function of both number of days observed, and number of user traces observed.
We see that stickiness predictions can be relatively accurate after observing only 10 days of data. The predictions improve as time passes, and having access to more user traces further increases predictive accuracy.

We now study the noise and prior covariance matrices $\bm{V}$ and $\bm{\Sigma}$, respectively.
We investigate how the variance of the sample reward, $\mathbb{V}[r \mid \bm{z}_{:t}, \bar{\bm{z}}]$, and the variance of the mean reward, $\mathbb{V}[\bar{r} \mid \bar{\bm{z}}_{:t}]$, are progressively explained away as $t$ increases, i.e., as we condition on more and more days observed.
Normalizing the $t$th conditional variance by the total (unconditional) variance, we obtain the fraction of total variance explained by the first $t$ intermediate outcomes.
Technical details are provided in Appendix~\ref{app:covar}, alongside visualizations of the covariance matrices as heatmaps.

\begin{figure}[t]
  \centering
  \includegraphics{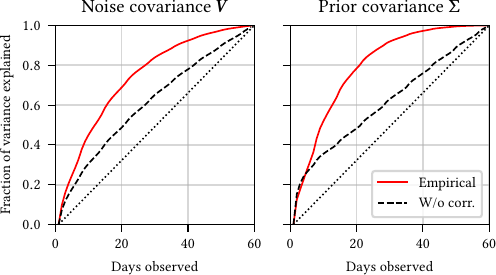}
  \caption{Explained variance as a function of days of activity data observed.}
  \Description{Plots visualizing the fraction of variance in user activity that is explained by the first $t$ elements of a covariance matrix.
  Left: noise covariance matrix $\bm{V}$. Right: prior covariance matrix $\bm{\Sigma}$.}
  \label{fig:varexplained}
\end{figure}

In Figure \ref{fig:varexplained} (left), we look at the noise covariance $\bm{V}$.
The diagonal straight line represents a hypothetical scenario where daily activity indicators $\bm{z}$ are distributed independently and identically around $\bar{\bm{z}}$, resulting in us gaining a constant amount of information about $r$ for each additional day of observed data.
The \emph{empirical} line corresponds to the actual covariance matrix learned by our approach.
We can see that around 10 days worth of data is sufficient to capture over 50\% of the aleatoric uncertainty in the reward $r$.
There are two factors that account for this.
The first is that, as time progresses, user activity reduces, so the variance is larger early on in the 60-day window;
This would be the case even if activity was entirely independent across days.
The second and more interesting factor, is that activity is correlated across days, therefore knowledge of activity up to a given day allows us to predict future activity.
The \emph{uncorrelated} line corresponds to the hypothetical case where the diagonal of the covariance matrix matches that of $\bm{V}$, but there is no correlation (i.e., off-diagonal elements of the matrix are set to zero).
The gap between the empirical and uncorrelated curves illustrates how much information we gain by exploiting the fact that past activity is predictive of future activity.

Similarly, in Figure~\ref{fig:varexplained} (right), we look at the prior covariance $\bm{\Sigma}$.
Intuitively, this lets us explore how much of the variance of $\bar{r}$ would be explained if we were to observe the first $t$ elements of each of a set of $M$ independent sample traces, as $M \to \infty$.
We can draw similar conclusions from this plot as to those mentioned in the context of $\bm{V}$.
However the trend is even more stark here, as 50\% of variance is explained by just eight days of data, and 95\% of the variance is explained within a month.

\subsection{Sequential Decision-Making Task}
\label{sec:evalbandit}

\begin{figure*}[t]
  \centering
  \includegraphics{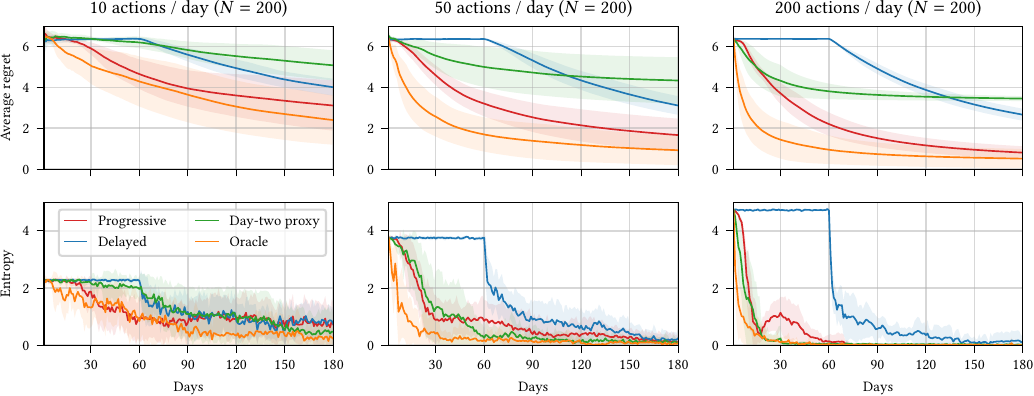}
  \caption{Average per-step regret and entropy of set of actions taken at each round, for $N = 200$ podcast shows.}
  \Description{Top row: plot of how the per-step regret incurred by the impatient bandit algorithm progresses across 180 rounds of bandit evaluation, along with three baselines (delayed, day-one and oracle feedback), in the scenario where we have a library of 200 podcast shows.
  Bottom row: plot of how the entropy of the actions taken by the impatient bandit algorithm progresses across 180 rounds of bandit evaluation, along with three baselines (delayed, day-two and oracle feedback), in the scenario where we have a library of 200 podcast shows.}
  \label{fig:bandit200}
\end{figure*}

We now turn our focus to the evaluation of the impatient bandit algorithm, and to the comparison of its empirical performance with competing approaches.
The way observed feedback is used is one of the main points of differentiation between the approaches we consider.
As such, we refer to our approach as \emph{progressive}, since we makes use of all observations as they are revealed over time.
We contrast the performance of our approach to three baselines.
\begin{description}
\item[Delayed.] The case in which we solely receive full observations, $\Delta$ days after an action is taken is referred to as the \emph{delayed} feedback baseline.
This naive approach does not attempt to take advantage of intermediate outcomes.

\item[Day-two proxy.] We treat the second day of activity as a proxy for stickiness and discard all subsequent information.
This baseline captures an intuitive outcome that is clearly related to the goal of maximizing habitual engagement: Does the user return to the show the day after discovering it? 
This baseline is representative of short-term proxies widely used in recommender systems, such as the click-through-rate, the dwell time, or the conversion rate \citep{bogina2017incorporating, dedieu2018hierarchical, lalmas2014measuring}.

\item[Oracle.] Finally, we include an \emph{oracle} baseline, which assumes that the full 60-day activity trace is received immediately after an action is taken.
This is clearly unrealistic, but it is useful to include as it provides an upper-bound on the performance of any model.
\end{description}
These baselines have been chosen to illustrate the benefits of incorporating progressive feedback into a bandit algorithm, and the effectiveness of our approach in making use of this intermediate information.
We use Thompson sampling for all of our baselines to ensure that any performance differences are due to the manner in which feedback is being considered, rather than to the relative strengths and weaknesses of different families of bandit algorithms.
For similar reasons, we do not compare to any works that study other aspects of recommendation unrelated to this study, such as personalization.

To mimic a realistic deployment setting in which the prior would be computed using data from the past, we compute our prior using the training data, and then run our algorithm on the unseen evaluation dataset.
A single prior is computed using all available traces from all 200 shows in our training set, this is then used for all of the experiments in this section.
We run the bandit for 180 rounds (corresponding to approximately 6 months), repeating each experiment 10 times to generate confidence intervals for the average regret.
Three different experimental setups are considered, with varying numbers of actions taken per day.

\subsubsection{Results}

Figure~\ref{fig:bandit200} (top row) visualizes the average per-step regret for each of these experimental settings, which ideally should tend to zero as $t \to \infty$.
Across all of the experiments, the performance of the delayed approach is poor as it is forced to make uninformed decisions for the first $\Delta$ rounds of evaluation due to the inherent delay in feedback being received.
Additionally, the oracle, as expected, outperforms the other approaches due to the unrealistic amount of information it has access to.
The day-two proxy approach performs well at first, comparably to our approach across the initial month of evaluation, but past this stage the limitations of optimizing for this proxy become clear.
The proxy is not well aligned, and the per-step regret rapidly plateaus.

Our progressive approach exhibits superior performance compared to the competing delayed and day-two proxy approaches; in fact, the performance of our approach is closer to that of the oracle.
As we increase the number of actions per round, we see a slight reduction in per-step regret across all approaches.

Figure~\ref{fig:bandit200} (bottom row) provides an alternative perspective on the outcome of these experiments, visualizing the entropy of the set of actions taken at each round.
Should a bandit converge on recommending a single show repeatedly at each round, the entropy would tend to zero.
The entropy plots show that, early on in the evaluation phase, our progressive algorithm tends to diversify across actions more than the oracle and day-two proxy.
The interpretation of this is that our approach is performing a broader exploration of the action space, a characteristic that can be very useful in a realistic, deployment setting, which we discuss below. 
Not only does Figure~\ref{fig:bandit200} let us compare the empirical performance of all four approaches, it also enables us to differentiate the effects of observational noise from the effects of delayed feedback.
For example, the large gap in per-step regret between the oracle and delayed approaches is entirely due to the delay in feedback, as both approaches receive full user traces of length $\Delta$.
On the other hand, the gap in per-step regret between the oracle and day-two proxy approaches is due to the fact that the second day of activity is a noisy proxy for the true stickiness, thus the day-two proxy approach tends to rapidly converge on a small subset of sub-optimal shows (this can be seen from its entropy, which quickly approaches zero).

\begin{figure*}
  \centering
  \includegraphics{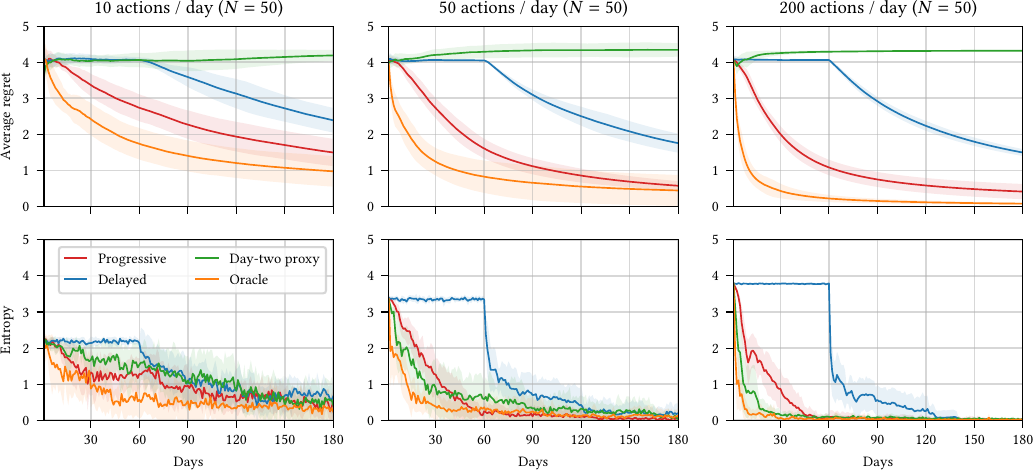}
  \caption{Average per-step regret and entropy of set of actions taken at each round, for $N=50$ podcast shows.}
  \Description{Top row: plot of how the per-step regret incurred by the impatient bandit algorithm progresses across 180 rounds of bandit evaluation, along with three baselines (delayed, day-one and oracle feedback), in the scenario where we have a library of 50 podcast shows..
  Bottom row: plot of how the entropy of the actions taken by the impatient bandit algorithm progresses across 180 rounds of bandit evaluation, along with three baselines (delayed, day-one and oracle feedback), in the scenario where we have a library of 50 podcast shows..}
  \label{fig:bandit50}
\end{figure*}

In Figure \ref{fig:bandit50}, we present additional results for a scenario in which we have a smaller action space, consisting of a subset of \num{50} shows sampled from the original evaluation dataset discussed previously.
This is clearly a simpler problem setting, as evidenced by the fact that all of the approaches tend more quickly to lower values of average regret in this case except for the day-two proxy feedback.
Besides this observation, the results follow largely similar trends to those seen in Figure \ref{fig:bandit200}.

\begin{figure}[t]
  \centering
  \includegraphics{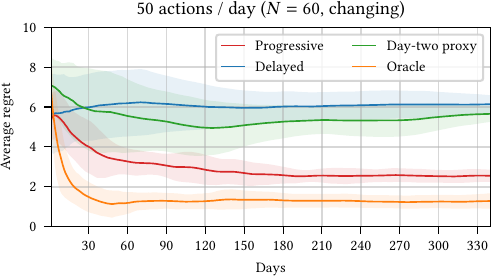}
  \caption{Average per-step regret in a scenario where we have a \emph{changing} set of 60 podcast shows.}
  \label{fig:changing}
  \Description{Plot of how the per-step regret incurred by the impatient bandit algorithm progresses across 180 rounds of bandit evaluation, along with three baselines (delayed, day-one and oracle feedback), in the scenario where we have a changing library of 60 podcast shows.}
\end{figure}

\paragraph{Changing Show Set}
In addition to considering a static library of shows, we also briefly consider a setting where we have a library of shows that is constantly evolving over time.
Specifically, at each round, one randomly selected show is removed from the library and is replaced with a new show.
From the results shown in Figure~\ref{fig:changing}, we can see that our algorithm once again considerably outperforms the delayed and day-two proxy feedback schemes, even in this challenging setting where new content is constantly entering the system and exploration is always necessary.

%% file: 05-conclusion.tex
\section{Conclusion \& Future Work}
\label{sec:conclusion}

In this work we have introduced a new type of bandit algorithm that efficiently optimizes for delayed rewards, assuming that intermediate outcomes correlated with the final reward are revealed progressively over time.
This is achieved by way of a meta-learning approach.
We begin by learning the parameters of a Bayesian filter by using historical data from a related but distinct problem.
Then, we combine this probabilistic reward model with Thompson sampling, effectively balancing exploration and exploitation.
The key to our success is that the Bayesian filter is able to make accurate inferences on delayed rewards using intermediate outcomes.

We have evaluated our framework empirically on a podcast content exploration problem.
Using real-world platform data, experimental results show that our approach, which utilizes all available intermediate information to estimate a long-term reward, significantly outperforms approaches that only use short-term proxies or wait until the reward is available.

We have presented a non-personalized methodology for optimizing recommendations over an extended period of time.
A natural avenue for future work is extending this to a personalized setting.
Conceptually, we do not foresee any major difficulty.
In Appendix~\ref{app:contextual}, we sketch an contextual extension of our Bayesian filter that conditions beliefs on user embeddings.
Another avenue of research would be to build a theoretical understanding of the favorable empirical performance observed in practical applications.
Can we formally characterize the benefits of progressive feedback over delayed rewards in terms of the average regret?

Finally, we would like to emphasize that the general framework we present can also benefit other application domains, beyond recommendations on online content platforms.
For example, we believe our algorithm could be used to allocate resources in hyperparameter optimisation problems~\citep{li2017hyperband} by identifying more or less promising hyperparameter configurations in the early stages of training from an array of intermediate validation metrics.
This could significantly reduce the computational cost of training large models and its environmental impact, which has become a major concern in the ML community in recent years~\citep{lacoste2019quantifying}.

\begin{acks}
We thank the anonymous reviewers for their constructive feedback, which greatly contributed to improving this paper.
\end{acks}

%% file: 0A-model.tex
\section{Comments on Reward Model}
\label{app:model}

In Section~\ref{sec:training}, we mention that if the definition of the problem at hand does not directly imply a linear relation between a given set of intermediate observations and a long-term reward of interest, one might try to learn a model of target outcomes $y$ by solving a regression problem
\begin{align*}
\textstyle
\Argmin_{w} \sum_{(\bm{z}, y) \in \mathcal{D}} (y - \bm{w}^\top \bm{z})^2
\end{align*}
on some historical data $\mathcal{D}$.
In general, the reward $r = \bm{w}^\top \bm{z}$ will no longer be identical to the true long-term target $y$.

In this case, the reward can be thought of as a \emph{surrogate index}, as defined in \citet{athey2019surrogate}.
Provided that several assumptions hold, this approach is principled.
Among others, $y$ needs to be independent of the selected action $a$ given $r$ (a.k.a. the surrogacy assumption).
This assumption requires $\bm{z}$ to contain sufficient information on $a$ as it relates to $y$.
Furthermore, the reward $r = \bm{w}^\top \bm{z}$ learned on historical training data should generalize to data coming in during evaluation (a.k.a. the comparability assumption).
In practice, it might be important to test these assumptions empirically.

\subsection{Non-Linear Extension}

The assumption that the reward $r$ is linear in the trace $\bm{z}$ is not as restrictive as it might appear at first sight.
It is easy to extend the model to capture non-linear relationships between $\bm{z}$ and $r$, while staying in the same linear-Gaussian framework that we rely on throughout Section~\ref{sec:method}.

As a concrete example, consider a reward $r$ that depends on $\bm{z}\in \mathbb{R}^2$ in a non-linear way, for example $r = z_1^2 - 3z_2 + 6z_1 z_2$.
We can augment the trace into a new vector $\bm{z}' = (z_1', z_2', z_3', z_4', z_5') = (z_1, z_2, z_1^2, z_2^2, z_1 \cdot z_2)$.
Now, we can represent any quadratic relationship between  $\bm{z}$ and $r$ as a linear relationship between $\bm{z}'$ and $r$.
In particular, our example yields $r = \bm{w}^\top \bm{z}'$ with $\bm{w} = (0, -3, 1, 0, 6)$.
By instantiating the reward model over $\bm{z}'$ instead of $\bm{z}$, we can thus model non-linear (quadratic) relations between intermediate outcomes and long-term reward.
This idea can be extended to higher-order polynomials or (perhaps better) to regression splines~\citep{hastie2009elements}, and capture non-linear relationships in a flexible way.

%% file: 0B-covar.tex
\section{Covariance Visualizations}
\label{app:covar}

\begin{figure}[t]
  \centering
  \includegraphics{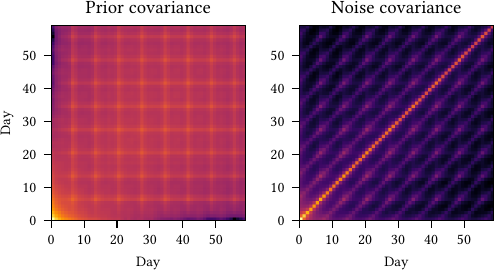}
  \caption{A visualization of the prior and noise covariance matrices.}
  \Description{A visualization of the prior and noise covariance matrices computed using our training dataset.}
  \label{fig:covmats}
\end{figure}

In Figure \ref{fig:covmats}, we show visualizations of the prior and noise covariance matrices $\bm{V}$ and $\bm{\Sigma}$ obtained by training the model on the data described in Section \ref{sec:datadesc}.
In the prior covariance matrix we see a clear weekly trend, and whilst the entries around the first few days of activity dominate, there is still a rich covariance structure across the whole 60-day period.
From the noise covariance, we can conclude that the daily observations are clearly not independent, but there is still a significant degree of day-to-day variability which is not explained.

\paragraph{Technical Details on Figure~\ref{fig:varexplained}}

This figure provides an additional perspective on the matrices $\bm{V}$ and $\bm{\Sigma}$.
On Figure~\ref{fig:varexplained} (left) we consider the noise covariance matrix $\bm{V}$.
We have
\begin{align*}
\mathbb{V}[r \mid \bm{z}_{:t}, \bar{\bm{z}}]
    &= \mathbb{V}[\bm{1}^\top \bm{z} \mid \bm{z}_{:t}, \bar{\bm{z}}]
    = \bm{1}^\top \tilde{\bm{V}}_t \bm{1} \doteq \tilde{v}^2_t, \\
\tilde{\bm{V}}_t
    &= \bm{V}_{t+1:,t+1:} - \bm{V}_{t+1:,:t} \bm{V}_{:t,:t}^{-1}\bm{V}_{:t,t+1:},
\end{align*}
where the last equality makes use of standard Gaussian identities~\citep[Section A.2]{rasmussen2006gaussian}.
We normalize the conditional variance by the total (unconditional) variance to obtain the fraction of total variance explained by the first $t$ intermediate outcomes, i.e., we report $\tilde{v}^2_t / \tilde{v}^2_0$.

On Figure~\ref{fig:varexplained} (right), we proceed similarly for the prior covariance matrix $\bm{\Sigma}$, computing
\begin{align*}
\mathbb{V}[\bar{r} \mid \bar{\bm{z}}_{:t}]
    &= \mathbb{V}[\bm{1}^\top \bar{\bm{z}} \mid \bar{\bm{z}}_{:t}]
    = \bm{1}^\top \tilde{\bm{\Sigma}}_t \bm{1} \doteq \tilde{\sigma}^2_t, \\
\tilde{\bm{\Sigma}}_t
    &= \bm{\Sigma}_{t+1:,t+1:} - \bm{\Sigma}_{t+1:,:t} \bm{\Sigma}_{:t,:t}^{-1}\bm{\Sigma}_{:t,t+1:},
\end{align*}
and we report the normalized value $\tilde{\sigma}^2_t / \tilde{\sigma}^2_0$.

%% file: 0C-contextual.tex
\section{Contextual Extension}
\label{app:contextual}

Our methodology can be extended to the contextual setting, and we briefly sketch this extension here.
For conciseness, let us consider the case of disjoint linear payoffs~\citep{li2010contextual}, and let us fix a single action and omit the subscript $a$.
Instead of modeling the $K$-dimensional average trace $\bar{\bm{z}}$, we now model a $(d \times K)$-dimensional matrix $\bm{\Theta}$.
We assume that the expected reward for selecting the action is $\bm{x}^\top \bm{\Theta} \bm{w}$, where $\bm{x}$ is a context vector (e.g., describing a user's preferences) that can change across rounds.
Intuitively, the $k$th column of $\bm{\Theta}$ describes coefficients of the $k$th context-dependent average intermediate outcome.

We can extend the Bayesian filter we describe in Section~\ref{sec:model} to model a belief over the random matrix $\bm{\Theta}$ instead of the random vector $\bar{\bm{z}}$.
This is achieved simply by vectorizing the matrix, that is, $\mathrm{vec}(\bm{\Theta}) \sim \mathcal{N}(\bm{\mu}, \bm{\Sigma})$, with $\bm{\mu}$ of dimension $dK$ and $\bm{\Sigma}$ of dimension $dK \times dK$.
We can condition the belief updates on the context using standard closed-form formulas.
For simplicity, consider a full trace $\bm{z}$ observed in context $\bm{x}$;
The posterior belief update is given by
\begin{align*}
\bm{A} &\gets \bm{\Sigma}(\bm{\Sigma} + \bm{x} \bm{x}^\top \otimes \bm{V})^{-1} \\
\bm{\mu}' &\gets \bm{\mu} + \bm{A}(\bm{x} \otimes \bm{z} - \bm{\mu}) \\
\bm{\Sigma}' &\gets \bm{\Sigma} + \bm{A}\bm{\Sigma},
\end{align*}
where $\otimes$ denotes the Kronecker product.

The simple training procedure described in Section~\ref{sec:training} cannot be easily extended to the contextual case, since the quantities involved in the averages are context-dependent.
Instead, we suggest using type-II maximum likelihood, a standard hyperparameter selection procedure.
We leave a detailed development of a contextual version of our approach for future work.

%% file: paper.bbl

\begin{thebibliography}{45}


\ifx \showCODEN    \undefined \def \showCODEN     #1{\unskip}     \fi
\ifx \showDOI      \undefined \def \showDOI       #1{#1}\fi
\ifx \showISBNx    \undefined \def \showISBNx     #1{\unskip}     \fi
\ifx \showISBNxiii \undefined \def \showISBNxiii  #1{\unskip}     \fi
\ifx \showISSN     \undefined \def \showISSN      #1{\unskip}     \fi
\ifx \showLCCN     \undefined \def \showLCCN      #1{\unskip}     \fi
\ifx \shownote     \undefined \def \shownote      #1{#1}          \fi
\ifx \showarticletitle \undefined \def \showarticletitle #1{#1}   \fi
\ifx \showURL      \undefined \def \showURL       {\relax}        \fi
\providecommand\bibfield[2]{#2}
\providecommand\bibinfo[2]{#2}
\providecommand\natexlab[1]{#1}
\providecommand\showeprint[2][]{arXiv:#2}

\bibitem[Agrawal and Goyal(2012)]%
        {agrawal2012analysis}
\bibfield{author}{\bibinfo{person}{Shipra Agrawal} {and} \bibinfo{person}{Navin
  Goyal}.} \bibinfo{year}{2012}\natexlab{}.
\newblock \showarticletitle{Analysis of {T}hompson sampling for the multi-armed
  bandit problem}. In \bibinfo{booktitle}{\emph{Conference on learning
  theory}}. JMLR Workshop and Conference Proceedings, \bibinfo{pages}{39--1}.
\newblock


\bibitem[Athey et~al\mbox{.}(2019)]%
        {athey2019surrogate}
\bibfield{author}{\bibinfo{person}{Susan Athey}, \bibinfo{person}{Raj Chetty},
  \bibinfo{person}{Guido~W Imbens}, {and} \bibinfo{person}{Hyunseung Kang}.}
  \bibinfo{year}{2019}\natexlab{}.
\newblock \bibinfo{booktitle}{\emph{The surrogate index: {C}ombining short-term
  proxies to estimate long-term treatment effects more rapidly and precisely}}.
\newblock \bibinfo{type}{{T}echnical {R}eport}. \bibinfo{institution}{National
  Bureau of Economic Research}.
\newblock


\bibitem[Aziz et~al\mbox{.}(2022)]%
        {aziz2022identifying}
\bibfield{author}{\bibinfo{person}{Maryam Aziz}, \bibinfo{person}{Jesse
  Anderton}, \bibinfo{person}{Kevin Jamieson}, \bibinfo{person}{Alice Wang},
  \bibinfo{person}{Hugues Bouchard}, {and} \bibinfo{person}{Javed Aslam}.}
  \bibinfo{year}{2022}\natexlab{}.
\newblock \showarticletitle{Identifying New Podcasts with High General Appeal
  Using a Pure Exploration Infinitely-Armed Bandit Strategy}. In
  \bibinfo{booktitle}{\emph{Proceedings of the 16th ACM Conference on
  Recommender Systems}}. \bibinfo{pages}{134--144}.
\newblock


\bibitem[Bastani et~al\mbox{.}(2022)]%
        {bastani2022meta}
\bibfield{author}{\bibinfo{person}{Hamsa Bastani}, \bibinfo{person}{David
  Simchi-Levi}, {and} \bibinfo{person}{Ruihao Zhu}.}
  \bibinfo{year}{2022}\natexlab{}.
\newblock \showarticletitle{Meta dynamic pricing: Transfer learning across
  experiments}.
\newblock \bibinfo{journal}{\emph{Management Science}} \bibinfo{volume}{68},
  \bibinfo{number}{3} (\bibinfo{year}{2022}), \bibinfo{pages}{1865--1881}.
\newblock


\bibitem[Basu et~al\mbox{.}(2021)]%
        {basu2021no}
\bibfield{author}{\bibinfo{person}{Soumya Basu}, \bibinfo{person}{Branislav
  Kveton}, \bibinfo{person}{Manzil Zaheer}, {and} \bibinfo{person}{Csaba
  Szepesv{\'a}ri}.} \bibinfo{year}{2021}\natexlab{}.
\newblock \showarticletitle{No regrets for learning the prior in bandits}.
\newblock \bibinfo{journal}{\emph{Advances in Neural Information Processing
  Systems}}  \bibinfo{volume}{34} (\bibinfo{year}{2021}),
  \bibinfo{pages}{28029--28041}.
\newblock


\bibitem[Bennett and Lanning(2007)]%
        {bennett2007netflix}
\bibfield{author}{\bibinfo{person}{James Bennett} {and} \bibinfo{person}{Stan
  Lanning}.} \bibinfo{year}{2007}\natexlab{}.
\newblock \showarticletitle{The {Netflix} Prize}. In
  \bibinfo{booktitle}{\emph{Proceedings of KDDCup '07}}. \bibinfo{address}{San
  Jose, CA, USA}.
\newblock


\bibitem[Bogina and Kuflik(2017)]%
        {bogina2017incorporating}
\bibfield{author}{\bibinfo{person}{Veronika Bogina} {and} \bibinfo{person}{Tsvi
  Kuflik}.} \bibinfo{year}{2017}\natexlab{}.
\newblock \showarticletitle{Incorporating Dwell Time in Session-Based
  Recommendations with Recurrent Neural Networks.}. In
  \bibinfo{booktitle}{\emph{RecTemp@ RecSys}}. \bibinfo{pages}{57--59}.
\newblock


\bibitem[Caria et~al\mbox{.}(2020)]%
        {caria2020adaptive}
\bibfield{author}{\bibinfo{person}{Stefano Caria}, \bibinfo{person}{Maximilian
  Kasy}, \bibinfo{person}{Simon Quinn}, \bibinfo{person}{Soha Shami},
  \bibinfo{person}{Alex Teytelboym}, {et~al\mbox{.}}}
  \bibinfo{year}{2020}\natexlab{}.
\newblock \showarticletitle{An adaptive targeted field experiment: {J}ob search
  assistance for refugees in {J}ordan}.
\newblock  (\bibinfo{year}{2020}).
\newblock


\bibitem[Chapelle and Li(2011)]%
        {chapelle2011empirical}
\bibfield{author}{\bibinfo{person}{Olivier Chapelle} {and}
  \bibinfo{person}{Lihong Li}.} \bibinfo{year}{2011}\natexlab{}.
\newblock \showarticletitle{An empirical evaluation of {T}hompson sampling}.
\newblock \bibinfo{journal}{\emph{Advances in Neural Information Processing
  Systems}}  \bibinfo{volume}{24} (\bibinfo{year}{2011}).
\newblock


\bibitem[Dedieu et~al\mbox{.}(2018)]%
        {dedieu2018hierarchical}
\bibfield{author}{\bibinfo{person}{Antoine Dedieu}, \bibinfo{person}{Rahul
  Mazumder}, \bibinfo{person}{Zhen Zhu}, {and} \bibinfo{person}{Hossein
  Vahabi}.} \bibinfo{year}{2018}\natexlab{}.
\newblock \showarticletitle{Hierarchical Modeling and Shrinkage for User
  Session Length Prediction in Media Streaming}. In
  \bibinfo{booktitle}{\emph{Proceedings of the 27th ACM International
  Conference on Information and Knowledge Management}}.
  \bibinfo{pages}{607--616}.
\newblock


\bibitem[Desautels et~al\mbox{.}(2014)]%
        {desautels2014parallelizing}
\bibfield{author}{\bibinfo{person}{Thomas Desautels}, \bibinfo{person}{Andreas
  Krause}, {and} \bibinfo{person}{Joel~W Burdick}.}
  \bibinfo{year}{2014}\natexlab{}.
\newblock \showarticletitle{Parallelizing exploration-exploitation tradeoffs in
  {G}aussian process bandit optimization}.
\newblock \bibinfo{journal}{\emph{Journal of Machine Learning Research}}
  \bibinfo{volume}{15} (\bibinfo{year}{2014}), \bibinfo{pages}{3873--3923}.
\newblock


\bibitem[Graepel et~al\mbox{.}(2010)]%
        {graepel2010web}
\bibfield{author}{\bibinfo{person}{Thore Graepel},
  \bibinfo{person}{Joaquin~Quinonero Candela}, \bibinfo{person}{Thomas
  Borchert}, {and} \bibinfo{person}{Ralf Herbrich}.}
  \bibinfo{year}{2010}\natexlab{}.
\newblock \showarticletitle{Web-scale {B}ayesian click-through rate prediction
  for sponsored search advertising in {M}icrosoft's {B}ing search engine}.
  Omnipress.
\newblock


\bibitem[Grover et~al\mbox{.}(2018)]%
        {grover2018best}
\bibfield{author}{\bibinfo{person}{Aditya Grover}, \bibinfo{person}{Todor
  Markov}, \bibinfo{person}{Peter Attia}, \bibinfo{person}{Norman Jin},
  \bibinfo{person}{Nicolas Perkins}, \bibinfo{person}{Bryan Cheong},
  \bibinfo{person}{Michael Chen}, \bibinfo{person}{Zi Yang},
  \bibinfo{person}{Stephen Harris}, \bibinfo{person}{William Chueh},
  {et~al\mbox{.}}} \bibinfo{year}{2018}\natexlab{}.
\newblock \showarticletitle{Best arm identification in multi-armed bandits with
  delayed feedback}. In \bibinfo{booktitle}{\emph{International Conference on
  Artificial Intelligence and Statistics}}. PMLR, \bibinfo{pages}{833--842}.
\newblock


\bibitem[Hastie et~al\mbox{.}(2009)]%
        {hastie2009elements}
\bibfield{author}{\bibinfo{person}{Trevor Hastie}, \bibinfo{person}{Robert
  Tibshirani}, {and} \bibinfo{person}{Jerome Friedman}.}
  \bibinfo{year}{2009}\natexlab{}.
\newblock \bibinfo{booktitle}{\emph{The Elements of Statistical Learning:
  {Data} Mining, Inference, and Prediction} (\bibinfo{edition}{second} ed.)}.
\newblock \bibinfo{publisher}{Springer}.
\newblock


\bibitem[Hohnhold et~al\mbox{.}(2015)]%
        {hohnhold2015focusing}
\bibfield{author}{\bibinfo{person}{Henning Hohnhold}, \bibinfo{person}{Deirdre
  O'Brien}, {and} \bibinfo{person}{Diane Tang}.}
  \bibinfo{year}{2015}\natexlab{}.
\newblock \showarticletitle{Focusing on the Long-Term: It's Good for Users and
  Business}. In \bibinfo{booktitle}{\emph{Proceedings of the 21th ACM SIGKDD
  International Conference on Knowledge Discovery and Data Mining}} (Sydney,
  NSW, Australia) \emph{(\bibinfo{series}{KDD '15})}.
  \bibinfo{publisher}{Association for Computing Machinery},
  \bibinfo{address}{New York, NY, USA}, \bibinfo{pages}{1849–1858}.
\newblock
\showISBNx{9781450336642}
\urldef\tempurl%
\url{https://doi.org/10.1145/2783258.2788583}
\showDOI{\tempurl}


\bibitem[Hong and Lalmas(2019)]%
        {hong2019tutorial}
\bibfield{author}{\bibinfo{person}{Liangjie Hong} {and} \bibinfo{person}{Mounia
  Lalmas}.} \bibinfo{year}{2019}\natexlab{}.
\newblock \showarticletitle{Tutorial on online user engagement: Metrics and
  optimization}. In \bibinfo{booktitle}{\emph{Companion Proceedings of The 2019
  World Wide Web Conference}}. \bibinfo{pages}{1303--1305}.
\newblock


\bibitem[Joulani et~al\mbox{.}(2013)]%
        {joulani2013online}
\bibfield{author}{\bibinfo{person}{Pooria Joulani}, \bibinfo{person}{Andras
  Gyorgy}, {and} \bibinfo{person}{Csaba Szepesv{\'a}ri}.}
  \bibinfo{year}{2013}\natexlab{}.
\newblock \showarticletitle{Online learning under delayed feedback}. In
  \bibinfo{booktitle}{\emph{International Conference on Machine Learning}}.
  PMLR, \bibinfo{pages}{1453--1461}.
\newblock


\bibitem[Kandasamy et~al\mbox{.}(2018)]%
        {kandasamy2018parallelised}
\bibfield{author}{\bibinfo{person}{Kirthevasan Kandasamy},
  \bibinfo{person}{Akshay Krishnamurthy}, \bibinfo{person}{Jeff Schneider},
  {and} \bibinfo{person}{Barnab{\'a}s P{\'o}czos}.}
  \bibinfo{year}{2018}\natexlab{}.
\newblock \showarticletitle{Parallelised {B}ayesian optimisation via {T}hompson
  sampling}. In \bibinfo{booktitle}{\emph{International Conference on
  Artificial Intelligence and Statistics}}. PMLR, \bibinfo{pages}{133--142}.
\newblock


\bibitem[Lacoste et~al\mbox{.}(2019)]%
        {lacoste2019quantifying}
\bibfield{author}{\bibinfo{person}{Alexandre Lacoste},
  \bibinfo{person}{Alexandra Luccioni}, \bibinfo{person}{Victor Schmidt}, {and}
  \bibinfo{person}{Thomas Dandres}.} \bibinfo{year}{2019}\natexlab{}.
\newblock \showarticletitle{Quantifying the carbon emissions of machine
  learning}.
\newblock \bibinfo{journal}{\emph{arXiv preprint arXiv:1910.09700}}
  (\bibinfo{year}{2019}).
\newblock


\bibitem[Lalmas et~al\mbox{.}(2014)]%
        {lalmas2014measuring}
\bibfield{author}{\bibinfo{person}{Mounia Lalmas}, \bibinfo{person}{Heather
  O'Brien}, {and} \bibinfo{person}{Elad Yom-Tov}.}
  \bibinfo{year}{2014}\natexlab{}.
\newblock \showarticletitle{Measuring user engagement}.
\newblock \bibinfo{journal}{\emph{Synthesis lectures on information concepts,
  retrieval, and services}} \bibinfo{volume}{6}, \bibinfo{number}{4}
  (\bibinfo{year}{2014}), \bibinfo{pages}{1--132}.
\newblock


\bibitem[Li et~al\mbox{.}(2010)]%
        {li2010contextual}
\bibfield{author}{\bibinfo{person}{Lihong Li}, \bibinfo{person}{Wei Chu},
  \bibinfo{person}{John Langford}, {and} \bibinfo{person}{Robert~E Schapire}.}
  \bibinfo{year}{2010}\natexlab{}.
\newblock \showarticletitle{A contextual-bandit approach to personalized news
  article recommendation}. In \bibinfo{booktitle}{\emph{Proceedings of the 19th
  International Conference on World Wide Web}}. \bibinfo{pages}{661--670}.
\newblock


\bibitem[Li et~al\mbox{.}(2017)]%
        {li2017hyperband}
\bibfield{author}{\bibinfo{person}{Lisha Li}, \bibinfo{person}{Kevin Jamieson},
  \bibinfo{person}{Giulia DeSalvo}, \bibinfo{person}{Afshin Rostamizadeh},
  {and} \bibinfo{person}{Ameet Talwalkar}.} \bibinfo{year}{2017}\natexlab{}.
\newblock \showarticletitle{{H}yperband: A novel bandit-based approach to
  hyperparameter optimization}.
\newblock \bibinfo{journal}{\emph{The Journal of Machine Learning Research}}
  \bibinfo{volume}{18}, \bibinfo{number}{1} (\bibinfo{year}{2017}),
  \bibinfo{pages}{6765--6816}.
\newblock


\bibitem[Mattos et~al\mbox{.}(2019)]%
        {mattos2019multi}
\bibfield{author}{\bibinfo{person}{David~Issa Mattos}, \bibinfo{person}{Jan
  Bosch}, {and} \bibinfo{person}{Helena~Holmstr{\"o}m Olsson}.}
  \bibinfo{year}{2019}\natexlab{}.
\newblock \showarticletitle{Multi-armed bandits in the wild: Pitfalls and
  strategies in online experiments}.
\newblock \bibinfo{journal}{\emph{Information and Software Technology}}
  \bibinfo{volume}{113} (\bibinfo{year}{2019}), \bibinfo{pages}{68--81}.
\newblock


\bibitem[Maystre et~al\mbox{.}(2023)]%
        {maystre2023optimizing}
\bibfield{author}{\bibinfo{person}{Lucas Maystre}, \bibinfo{person}{Dan Russo},
  {and} \bibinfo{person}{Yu Zhao}.} \bibinfo{year}{2023}\natexlab{}.
\newblock \bibinfo{title}{Optimizing Audio Recommendations for the Long-Term:
  {A} Reinforcement Learning Perspective}.  (\bibinfo{date}{Feb.}
  \bibinfo{year}{2023}).
\newblock
\newblock
\shownote{Preprint, \texttt{arXiv:2302.03561v2 [cs.LG]}}.


\bibitem[McInerney et~al\mbox{.}(2018)]%
        {mcinerney2018explore}
\bibfield{author}{\bibinfo{person}{James McInerney}, \bibinfo{person}{Benjamin
  Lacker}, \bibinfo{person}{Samantha Hansen}, \bibinfo{person}{Karl Higley},
  \bibinfo{person}{Hugues Bouchard}, \bibinfo{person}{Alois Gruson}, {and}
  \bibinfo{person}{Rishabh Mehrotra}.} \bibinfo{year}{2018}\natexlab{}.
\newblock \showarticletitle{Explore, Exploit, and Explain: {Personalizing}
  Explainable Recommendations with Nandits}. In
  \bibinfo{booktitle}{\emph{Proceedings of RecSys '18}}.
  \bibinfo{address}{Vancouver, BC, Canada}.
\newblock


\bibitem[Prentice(1989)]%
        {prentice1989surrogate}
\bibfield{author}{\bibinfo{person}{Ross~L Prentice}.}
  \bibinfo{year}{1989}\natexlab{}.
\newblock \showarticletitle{Surrogate endpoints in clinical trials: definition
  and operational criteria}.
\newblock \bibinfo{journal}{\emph{Statistics in Medicine}} \bibinfo{volume}{8},
  \bibinfo{number}{4} (\bibinfo{year}{1989}), \bibinfo{pages}{431--440}.
\newblock


\bibitem[Qin and Russo(2022)]%
        {qin2022adaptivity}
\bibfield{author}{\bibinfo{person}{Chao Qin} {and} \bibinfo{person}{Daniel
  Russo}.} \bibinfo{year}{2022}\natexlab{}.
\newblock \showarticletitle{Adaptivity and confounding in multi-armed bandit
  experiments}.
\newblock \bibinfo{journal}{\emph{arXiv preprint arXiv:2202.09036}}
  (\bibinfo{year}{2022}).
\newblock


\bibitem[Rasmussen et~al\mbox{.}(2006)]%
        {rasmussen2006gaussian}
\bibfield{author}{\bibinfo{person}{Carl~Edward Rasmussen},
  \bibinfo{person}{Christopher~KI Williams}, {et~al\mbox{.}}}
  \bibinfo{year}{2006}\natexlab{}.
\newblock \bibinfo{booktitle}{\emph{{G}aussian processes for machine
  learning}}. Vol.~\bibinfo{volume}{1}.
\newblock \bibinfo{publisher}{Springer}.
\newblock


\bibitem[Ricci et~al\mbox{.}(2015)]%
        {ricci2015recommender}
\bibfield{author}{\bibinfo{person}{F. Ricci}, \bibinfo{person}{L. Rokach},
  {and} \bibinfo{person}{B. Shapira}.} \bibinfo{year}{2015}\natexlab{}.
\newblock \bibinfo{booktitle}{\emph{Recommender Systems Handbook}
  (\bibinfo{edition}{second} ed.)}.
\newblock \bibinfo{publisher}{Springer}.
\newblock


\bibitem[Russo and Van~Roy(2016)]%
        {russo2016information}
\bibfield{author}{\bibinfo{person}{Daniel Russo} {and}
  \bibinfo{person}{Benjamin Van~Roy}.} \bibinfo{year}{2016}\natexlab{}.
\newblock \showarticletitle{An information-theoretic analysis of {Thompson}
  sampling}.
\newblock \bibinfo{journal}{\emph{The Journal of Machine Learning Research}}
  \bibinfo{volume}{17}, \bibinfo{number}{1} (\bibinfo{year}{2016}),
  \bibinfo{pages}{2442--2471}.
\newblock


\bibitem[Russo et~al\mbox{.}(2018)]%
        {russo2018tutorial}
\bibfield{author}{\bibinfo{person}{Daniel~J Russo}, \bibinfo{person}{Benjamin
  Van~Roy}, \bibinfo{person}{Abbas Kazerouni}, \bibinfo{person}{Ian Osband},
  \bibinfo{person}{Zheng Wen}, {et~al\mbox{.}}}
  \bibinfo{year}{2018}\natexlab{}.
\newblock \showarticletitle{A tutorial on {T}hompson sampling}.
\newblock \bibinfo{journal}{\emph{Foundations and Trends{\textregistered} in
  Machine Learning}} \bibinfo{volume}{11}, \bibinfo{number}{1}
  (\bibinfo{year}{2018}), \bibinfo{pages}{1--96}.
\newblock


\bibitem[Scott(2010)]%
        {scott2010modern}
\bibfield{author}{\bibinfo{person}{Steven~L Scott}.}
  \bibinfo{year}{2010}\natexlab{}.
\newblock \showarticletitle{A modern {B}ayesian look at the multi-armed
  bandit}.
\newblock \bibinfo{journal}{\emph{Applied Stochastic Models in Business and
  Industry}} \bibinfo{volume}{26}, \bibinfo{number}{6} (\bibinfo{year}{2010}),
  \bibinfo{pages}{639--658}.
\newblock


\bibitem[Simchowitz et~al\mbox{.}(2021)]%
        {simchowitz2021bayesian}
\bibfield{author}{\bibinfo{person}{Max Simchowitz},
  \bibinfo{person}{Christopher Tosh}, \bibinfo{person}{Akshay Krishnamurthy},
  \bibinfo{person}{Daniel~J Hsu}, \bibinfo{person}{Thodoris Lykouris},
  \bibinfo{person}{Miro Dudik}, {and} \bibinfo{person}{Robert~E Schapire}.}
  \bibinfo{year}{2021}\natexlab{}.
\newblock \showarticletitle{{B}ayesian decision-making under misspecified
  priors with applications to meta-learning}.
\newblock \bibinfo{journal}{\emph{Advances in Neural Information Processing
  Systems}}  \bibinfo{volume}{34} (\bibinfo{year}{2021}),
  \bibinfo{pages}{26382--26394}.
\newblock


\bibitem[Slivkins et~al\mbox{.}(2019)]%
        {slivkins2019introduction}
\bibfield{author}{\bibinfo{person}{Aleksandrs Slivkins} {et~al\mbox{.}}}
  \bibinfo{year}{2019}\natexlab{}.
\newblock \showarticletitle{Introduction to multi-armed bandits}.
\newblock \bibinfo{journal}{\emph{Foundations and Trends{\textregistered} in
  Machine Learning}} \bibinfo{volume}{12}, \bibinfo{number}{1-2}
  (\bibinfo{year}{2019}), \bibinfo{pages}{1--286}.
\newblock


\bibitem[Srinivas et~al\mbox{.}(2009)]%
        {srinivas2009gaussian}
\bibfield{author}{\bibinfo{person}{Niranjan Srinivas}, \bibinfo{person}{Andreas
  Krause}, \bibinfo{person}{Sham~M Kakade}, {and} \bibinfo{person}{Matthias
  Seeger}.} \bibinfo{year}{2009}\natexlab{}.
\newblock \showarticletitle{{G}aussian process optimization in the bandit
  setting: No regret and experimental design}.
\newblock \bibinfo{journal}{\emph{arXiv preprint arXiv:0912.3995}}
  (\bibinfo{year}{2009}).
\newblock


\bibitem[Thompson(1933)]%
        {thompson1933likelihood}
\bibfield{author}{\bibinfo{person}{William~R Thompson}.}
  \bibinfo{year}{1933}\natexlab{}.
\newblock \showarticletitle{On the likelihood that one unknown probability
  exceeds another in view of the evidence of two samples}.
\newblock \bibinfo{journal}{\emph{Biometrika}} \bibinfo{volume}{25},
  \bibinfo{number}{3-4} (\bibinfo{year}{1933}), \bibinfo{pages}{285--294}.
\newblock


\bibitem[Tran et~al\mbox{.}(2021)]%
        {tran2021recommender}
\bibfield{author}{\bibinfo{person}{Thi Ngoc~Trang Tran},
  \bibinfo{person}{Alexander Felfernig}, \bibinfo{person}{Christoph Trattner},
  {and} \bibinfo{person}{Andreas Holzinger}.} \bibinfo{year}{2021}\natexlab{}.
\newblock \showarticletitle{Recommender systems in the healthcare domain:
  {State-of-the-art} and research issues}.
\newblock \bibinfo{journal}{\emph{Journal of Intelligent Information Systems}}
  \bibinfo{volume}{57} (\bibinfo{year}{2021}), \bibinfo{pages}{171--201}.
\newblock


\bibitem[Verbert et~al\mbox{.}(2012)]%
        {verbert2012context}
\bibfield{author}{\bibinfo{person}{Katrien Verbert}, \bibinfo{person}{Nikos
  Manouselis}, \bibinfo{person}{Xavier Ochoa}, \bibinfo{person}{Martin
  Wolpers}, \bibinfo{person}{Hendrik Drachsler}, \bibinfo{person}{Ivana
  Bosnic}, {and} \bibinfo{person}{Erik Duval}.}
  \bibinfo{year}{2012}\natexlab{}.
\newblock \showarticletitle{Context-Aware Recommender Systems for Learning: {A}
  Survey and Future Challenges}.
\newblock \bibinfo{journal}{\emph{Journal of Intelligent Information Systems}}
  \bibinfo{volume}{5}, \bibinfo{number}{4} (\bibinfo{year}{2012}),
  \bibinfo{pages}{318--335}.
\newblock


\bibitem[Wu and Wager(2022a)]%
        {wu2022partial}
\bibfield{author}{\bibinfo{person}{Han Wu} {and} \bibinfo{person}{Stefan
  Wager}.} \bibinfo{year}{2022}\natexlab{a}.
\newblock \showarticletitle{Partial Likelihood Thompson Sampling}.
\newblock \bibinfo{journal}{\emph{arXiv preprint arXiv:2203.00820}}
  (\bibinfo{year}{2022}).
\newblock


\bibitem[Wu and Wager(2022b)]%
        {wu2022thompson}
\bibfield{author}{\bibinfo{person}{Han Wu} {and} \bibinfo{person}{Stefan
  Wager}.} \bibinfo{year}{2022}\natexlab{b}.
\newblock \showarticletitle{{T}hompson Sampling with Unrestricted Delays}.
\newblock \bibinfo{journal}{\emph{arXiv preprint arXiv:2202.12431}}
  (\bibinfo{year}{2022}).
\newblock


\bibitem[Wu et~al\mbox{.}(2017)]%
        {wu2017returning}
\bibfield{author}{\bibinfo{person}{Qingyun Wu}, \bibinfo{person}{Hongning
  Wang}, \bibinfo{person}{Liangjie Hong}, {and} \bibinfo{person}{Yue Shi}.}
  \bibinfo{year}{2017}\natexlab{}.
\newblock \showarticletitle{Returning is believing: Optimizing long-term user
  engagement in recommender systems}. In \bibinfo{booktitle}{\emph{Proceedings
  of the 2017 ACM on Conference on Information and Knowledge Management}}.
  \bibinfo{pages}{1927--1936}.
\newblock


\bibitem[Yang et~al\mbox{.}(2020)]%
        {yang2020targeting}
\bibfield{author}{\bibinfo{person}{Jeremy Yang}, \bibinfo{person}{Dean Eckles},
  \bibinfo{person}{Paramveer Dhillon}, {and} \bibinfo{person}{Sinan Aral}.}
  \bibinfo{year}{2020}\natexlab{}.
\newblock \showarticletitle{Targeting for long-term outcomes}.
\newblock \bibinfo{journal}{\emph{arXiv preprint arXiv:2010.15835}}
  (\bibinfo{year}{2020}).
\newblock


\bibitem[Zhao et~al\mbox{.}(2019)]%
        {zhao2019deep}
\bibfield{author}{\bibinfo{person}{Xiangyu Zhao}, \bibinfo{person}{Long Xia},
  \bibinfo{person}{Jiliang Tang}, {and} \bibinfo{person}{Dawei Yin}.}
  \bibinfo{year}{2019}\natexlab{}.
\newblock \showarticletitle{Deep reinforcement learning for search,
  recommendation, and online advertising: a survey}.
\newblock \bibinfo{journal}{\emph{ACM {SIGWEB} newsletter}}
  \bibinfo{number}{Spring} (\bibinfo{year}{2019}), \bibinfo{pages}{1--15}.
\newblock


\bibitem[Zheng et~al\mbox{.}(2018)]%
        {zheng2018drn}
\bibfield{author}{\bibinfo{person}{Guanjie Zheng}, \bibinfo{person}{Fuzheng
  Zhang}, \bibinfo{person}{Zihan Zheng}, \bibinfo{person}{Yang Xiang},
  \bibinfo{person}{Nicholas~Jing Yuan}, \bibinfo{person}{Xing Xie}, {and}
  \bibinfo{person}{Zhenhui Li}.} \bibinfo{year}{2018}\natexlab{}.
\newblock \showarticletitle{{DRN}: A deep reinforcement learning framework for
  news recommendation}. In \bibinfo{booktitle}{\emph{Proceedings of the 2018
  World Wide Web Conference}}. \bibinfo{pages}{167--176}.
\newblock


\bibitem[Zou et~al\mbox{.}(2019)]%
        {zou2019reinforcement}
\bibfield{author}{\bibinfo{person}{Lixin Zou}, \bibinfo{person}{Long Xia},
  \bibinfo{person}{Zhuoye Ding}, \bibinfo{person}{Jiaxing Song},
  \bibinfo{person}{Weidong Liu}, {and} \bibinfo{person}{Dawei Yin}.}
  \bibinfo{year}{2019}\natexlab{}.
\newblock \showarticletitle{Reinforcement learning to optimize long-term user
  engagement in recommender systems}. In \bibinfo{booktitle}{\emph{Proceedings
  of the 25th ACM SIGKDD International Conference on Knowledge Discovery \&
  Data Mining}}. \bibinfo{pages}{2810--2818}.
\newblock


\end{thebibliography}
